\documentclass{isprs} 
\usepackage{amsmath}
\usepackage{amssymb}
\usepackage{comment}
\usepackage{subfigure}
\usepackage{setspace}
\usepackage{geometry} 
\usepackage{epstopdf}
\usepackage[labelsep=period]{caption}  
\usepackage[british]{babel} 
\usepackage[hang]{footmisc}

\usepackage{xcolor}

\usepackage{amsmath}

\newcommand{\Exp}{\mathrm{Exp}}
\newcommand{\Log}{\mathrm{Log}}
\newcommand{\mbf}[1]{\mathbf{#1}}

\usepackage{xspace}
\makeatletter
\DeclareRobustCommand\onedot{\futurelet\@let@token\@onedot}
\def\@onedot{\ifx\@let@token.\else.\null\fi\xspace}

\def\eg{\emph{e.g}\onedot} 
\def\ie{\emph{i.e}\onedot}

\def\etal{\emph{et al}\onedot}
\makeatother

\begin{document}


\title{IMU Propagation as Preintegration}

\author{
Jianzhu Huai\textsuperscript{1}
}

\address{
\textsuperscript{1 }State Key Lab of Info Engineering in Surveying, Mapping and Remote Sensing, \\Wuhan University, Wuhan, Hubei China - jianzhu.huai@whu.edu.cn
}


\abstract{
IMU preintegration is widely used in factor-graph-based visual--inertial, lidar--inertial, and radar--inertial state estimation, yet it is often treated as a specialized implementation separate from conventional IMU propagation.
This note shows that IMU preintegration and propagation are equivalent realizations of the same underlying computation. We present a convention-agnostic view in which 
the preintegrated measurement, bias Jacobians, and covariance can be obtained by wrapping an existing IMU propagation routine,
while a preintegration module can conversely recover state-transition matrices and propagated covariances.
This perspective simplifies the reuse of existing propagation code, supports translation across different error-state definitions, and provides practical consistency checks for preintegration implementations.
Experiments with random IMU sequences demonstrate close agreement between an RK4-based propagation implementation and GTSAM's tangent and manifold preintegration modules in the recovered Jacobians, covariances, and transition matrices.
}

\keywords{IMU Propagation, IMU Preintegration, Multisensor Fusion}
\maketitle

\section{Introduction}\label{sec:intro}

\sloppy

IMU preintegration \cite{delamaEquivariant2025} has become a standard component in factor-graph-based state estimation, including visual--inertial \cite{huaiConsistentRightinvariantFixedlag2021}, lidar--inertial \cite{liHighperformanceSolidstatelidarinertialOdometry2021}, and radar--inertial \cite{huai2026ramba} odometry and mapping systems. Compared with direct IMU propagation, preintegration offers a convenient way to summarize many high-rate IMU measurements between two key states into a single factor that can be reused during iterative optimization. This makes it especially attractive in graph-based estimation, where the same IMU interval may be evaluated many times.

Despite its popularity, IMU preintegration is often perceived as requiring a dedicated implementation that is separate from conventional IMU propagation. In practice, however, many codebases already contain a reliable propagation module, often tied to a particular state or error-state definition. This raises two practical questions. First, does adopting IMU preintegration require reimplementing the IMU model from scratch? Second, how can one validate that a preintegration implementation, especially its bias Jacobians and covariance, is correct?

Classic works on IMU preintegration \cite{luptonVisualinertialaidedNavigationHighdynamic2012,forsterOnmanifoldPreintegrationRealtime2017} provide elegant derivations. However, these formulations are usually presented as standalone constructions under specific perturbation conventions, which can obscure their relation to the more familiar IMU propagation used in filtering and navigation code. As a result, porting preintegration across different error-state definitions, or reusing an existing propagation implementation, may appear more complicated than it actually is.

This note shows that IMU preintegration and IMU propagation can be viewed as two equivalent realizations of the same underlying computation. We first describe both in a way that is not tied to a particular perturbation convention. We then show that the preintegrated measurement, its Jacobian with respect to the initial IMU bias, and its covariance can all be obtained by wrapping an existing IMU propagation routine. Conversely, a preintegration module can be used to recover state-transition matrices and propagated covariances. This view also clarifies how to adapt preintegration across different error-state definitions without re-deriving bias Jacobians and residual covariances from scratch.

We validate the analysis by converting an RK4-based IMU propagation implementation to and from the GTSAM preintegration modules. In experiments with random IMU sequences, the recovered Jacobians, covariances, and transition matrices closely match those produced by GTSAM's tangent and manifold preintegration. These results suggest that a robust propagation implementation can serve both as a simple path to preintegration and as a practical reference for validating preintegration code.

\section{Background}

\subsection{Propagation}

The system state $\mathbf{x}$ typically includes the IMU position $\mathbf{p}_s$, velocity $\mathbf{v}_s$, and orientation $\mathbf{R}_{ws}$, as well as the gyroscope and accelerometer biases $\mathbf{b} = [\mathbf{b}_g, \mathbf{b}_a]$. In some formulations, the gravity vector $\mathbf{g}$ is also included in the state. A perturbation, or error state, is defined by
\begin{equation}
\mathbf{x} = \widehat{\mathbf{x}} \boxplus^p \delta \mathbf{x}^p,
\end{equation}
where the perturbation may be applied on either the left or the right of the estimate $\widehat{\mathbf{x}}$. The superscript $^p$ indicates that $\boxplus$ is defined for propagation, and will be omitted when no ambiguity arises.

Given 6-axis IMU measurements $\mathbf{z}_k = [\boldsymbol{\omega}_k^m, \mathbf{a}_k^m]$ at times $t_k$, $k \in [s,e]$, IMU propagation can be summarized as
\begin{align}
\mathrm{Prop}: [\mathbf{x}_s,\ \mathbf{I},\ \Sigma_s,\ \mathbf{z}_{s:e}]
& \rightarrow
[\mathbf{x}_{e|s},\ \Phi(t_e,t_s),\ \Sigma_{e|s}], \label{eq:prop}
\\
\Phi(t_e,t_s) &\doteq \boldsymbol{\Phi}_{e|s} \doteq \frac{\partial \delta\mbf x_{e|s}}{\partial \delta \mbf x_s},
\end{align}
where $\mathbf{x}_{e|s}$ is the predicted state at time $t_e$, $\Phi(t_e,t_s)$ is the state-transition matrix from $t_s$ to $t_e$, and $\Sigma_s$ and $\Sigma_{e|s}$ are the covariances of the error states $\delta \mbf x_s$ and $\delta \mbf x_{e|s}$ at $t_s$ and $t_e$, respectively. For clarity, we use the subscripts $s$ and $e$ to denote the start and end times.

The continuous-time system model with control input $\mathbf{u}$ and Gaussian noise $\mathbf{n}\sim\mathcal{N}(\mathbf{0},\mathbf{Q}\delta(t))$ is
\begin{align}
\frac{d}{dt}\mathbf{x} &= f(\mathbf{x}, \mathbf{u}, \mathbf{n}), \label{eq:cont-state}\\
\frac{d}{dt}\delta\mathbf{x} &=
\mathbf{F}(\mathbf{x}, \mathbf{u})\,\delta\mathbf{x}
+ \mathbf{G}(t)\,\mathbf{n}. \label{eq:cont-error}
\end{align}
After discretization over $dt = t_k - t_{k-1}$, with discrete noise $\mathbf{w}_k \sim \mathcal{N}(\mathbf{0},\mathbf{Q}/dt)$, we obtain
\begin{align}
\mathbf{x}_k &= g(\mathbf{x}_{k-1}, \mathbf{u}_k, \mathbf{w}_k), \\
\delta \mathbf{x}_{k} &=
\Phi(t_k,t_{k-1})\,\delta \mathbf{x}_{k-1}
+ \mathbf{B}_{k}\,\mathbf{w}_{k},
\end{align}
where
\[
\Phi(t_k,t_{k-1}) = \exp\!\big(\mathbf{F}(\mathbf{x},\mathbf{u})\,dt\big)
\approx
\mathbf{I}
+ \mathbf{F}dt
+ \tfrac{1}{2}\mathbf{F}^2 dt^2.
\]
The full transition matrix is
\[
\Phi(t_e,t_s) = \Phi(t_e,t_{e-1}) \cdots \Phi(t_{s+1},t_s).
\]

State propagation integrates \eqref{eq:cont-state}, for example by using a fourth-order Runge--Kutta (RK4) method. The covariance is propagated as
\begin{equation}
\begin{aligned}
\Sigma_k &=
\Phi(t_k,t_{k-1})\,\Sigma_{k-1}\,\Phi^\top(t_k,t_{k-1}) \\
&\quad
+ \int_{t_{k-1}}^{t_k}
\Phi(t_k,\tau)\,\mathbf{G}(\tau)\mathbf{Q}\mathbf{G}^\top(\tau)\,\Phi^\top(t_k,\tau)\, d\tau \\
&\approx
\Phi(t_k,t_{k-1})\,\Sigma_{k-1}\,\Phi^\top(t_k,t_{k-1}) \\
&\quad
+ \tfrac{1}{2}\Big[
\Phi(t_k,t_{k-1})\,\mathbf{G}(t_{k-1})\,\mathbf{Q}\,\mathbf{G}^\top(t_{k-1})\,\Phi^\top(t_k,t_{k-1}) \\
&\qquad\quad
+ \mathbf{G}(t_k)\,\mathbf{Q}\,\mathbf{G}^\top(t_k)
\Big].
\end{aligned}
\end{equation}

\subsection{Preintegration}
The preintegrated inertial measurement, defined independently of the initial navigation state, is
\begin{align}
\Delta \mathbf{R}
&\doteq \Delta \mathbf{R}_{s e}
\doteq \prod_{k=s+1}^{e}
\Exp\!\left(\int_{t_{k-1}}^{t_k} \boldsymbol{\omega}(t)\,dt \right),
\label{subeq:preint-theta} \\
\Delta \mathbf{p}
&\doteq \Delta \mathbf{p}^{s}_{e}
\doteq \int_{t_s}^{t_e} \int_{t_s}^{\tau}
\mathbf{R}^{s}_{t}\,\mathbf{a}(t)\,dt\, d\tau,
\label{subeq:preint-p} \\
\Delta \mathbf{v}
&\doteq \Delta \mathbf{v}^{s}_{e}
\doteq \int_{t_s}^{t_e} \mathbf{R}^{s}_{t}\,\mathbf{a}(t)\,dt, \label{subeq:preint-v} \\
\Delta \mbf b &= \int_{t_s}^{t_e} n_{b} dt, \label{subeq:preint-b} \\
\Delta t &= t_e - t_s. \label{subeq:preint-t}
\end{align}
Here, $\mathbf{a}$ and $\boldsymbol{\omega}$ denote the true IMU specific force and angular velocity, free of IMU imperfections:
\[
\mathbf{a} = \mathbf{a}_m - \mathbf{b}_a - \mathbf{n}_a,\qquad
\boldsymbol{\omega} = \boldsymbol{\omega}_m - \mathbf{b}_g - \mathbf{n}_g.
\]
The operator $\Exp(\cdot)$ maps a rotation vector to a rotation matrix via
$\Exp(\boldsymbol{\phi}) = \exp([\boldsymbol{\phi}]_\times)$, and
$\Log(\cdot)$ denotes its inverse.

We also define the error state $\delta \mbf x_{se}$ of the preintegrated increment $\Delta \mbf x$ by
\begin{equation}
\Delta \mbf x = \Delta \widehat{\mbf x} \boxplus^g \delta \mbf x_{se}^g \in \mathbb{R}^{15},
\end{equation}
where the superscript $^g$ indicates that $\boxplus$ is defined for preintegration and is omitted when clear from context.

The preintegration operation can be summarized as
\begin{equation}
\mathrm{Preint}:\ 
[\mathbf{b}_s, \mathbf{z}_{s:e}, \mbf q]
\ \rightarrow\
[\Delta\mathbf{x}_{s e},\ \mathbf{J}_{b_s},\ \Sigma_{\Delta\mathbf{x}}],
\end{equation}
where
\begin{equation}
\mathbf{J}_{b_s}
=\frac{\partial\,\delta\mathbf{x}_{s e}}{\partial\,\delta\mathbf{b}_s} \in \mathbb{R}^{15\times 6}, \qquad
\Sigma_{\Delta\mathbf{x}} =\mathrm{cov}(\delta\mathbf{x}_{s e}) \in \mathbb{R}^{15\times 15},
\end{equation}
and $\mbf q$ denotes the IMU noise parameters, \eg, measurement noise densities and bias random-walk densities. Let $\mathbf{J}_{b_s}^n$ denote the first nine rows of $\mathbf{J}_{b_s}$. The last six rows, corresponding to the propagated biases, are trivially zero since the bias change over $\Delta t$ is unrelated to the bias error at $t_s$. The covariance $\Sigma_{\Delta\mathbf{x}}$ captures the uncertainty accumulated over the interval $[t_s,t_e]$ due to gyroscope and accelerometer noise.

Using these increments, the predicted IMU state is obtained with $\mbf f(\mbf x_s, \Delta \mbf x_{se}(\mbf b_s))$:
\begin{align}
\mathbf{R}_{w,e|s}
&\doteq \mathbf{R}_{w s}\,\Delta \mathbf{R},
\label{subeq:prop-by-int-rot} \\
\mathbf{p}_{e|s}
&\doteq \mathbf{p}_{s}
+ \mathbf{v}_{s}\,\Delta t
+ \tfrac{1}{2}\mathbf{g}\,\Delta t^{2}
+ \mathbf{R}_{w s}\,\Delta \mathbf{p},
\label{subeq:prop-by-int-p-state} \\
\mathbf{v}_{e|s}
&\doteq \mathbf{v}_{s}
+ \mathbf{R}_{w s}\,\Delta \mathbf{v}
+ \mathbf{g}\,\Delta t,
\label{subeq:prop-by-int-v} \\
\mathbf{b}_{e|s}
&\doteq \mathbf{b}_{s} + \Delta \mbf b_{se},
\label{subeq:prop-by-int-b}
\end{align}
where $\mbf g$ is the gravity in the world frame.
These equations form the bridge between preintegration and propagation.

\section{Preintegration from and to Propagation}

\subsection{Preintegration by Propagation}
\label{sec:preint_by_prop}

Suppose that an IMU propagation module is already available and we would like to reuse it to implement the preintegrated measurement.
First, $\mathbf{J}_{b_s}$ can be obtained as
\begin{equation}
\mathbf{J}_{b_s}
=
\frac{\partial \delta \mbf x_{se}^g}{\partial \delta \mbf b_s^g}
=
\left(\frac{\partial \delta \mbf x_{e|s}^p}{\partial \delta \mbf x_{se}^g}\right)^{-1}
\frac{\partial \delta \mbf x_{e|s}^p}{\partial \delta \mbf b_s^p}
\frac{\partial \delta \mbf b_{s}^p}{\partial \delta \mbf b_s^g},
\end{equation}
where $\frac{\partial \delta \mbf x_{e|s}^p}{\partial \delta \mbf x_{se}^g}$ follows from the definitions of $\delta \mbf x_{e|s}^p$, $\delta \mbf x_{se}^g$, and the prediction equations \eqref{subeq:prop-by-int-rot}--\eqref{subeq:prop-by-int-b}, and
$\frac{\partial \delta \mbf b_{s}^p}{\partial \delta \mbf b_s^g}$ follows from the bias perturbation definitions.

Second, the covariance of the preintegrated measurement is obtained by propagation starting with zero covariance:
\begin{equation}
\Sigma_{\Delta\mathbf{x}}
=
\left(\frac{\partial \delta \mbf x_{e|s}^p}{\partial \delta \mbf x_{se}^g}\right)^{-1}
\mathrm{cov}(\delta \mbf x_{e|s}^p)
\left(\frac{\partial \delta \mbf x_{e|s}^p}{\partial \delta \mbf x_{se}^g}\right)^{-T}.
\end{equation}

These relations show that IMU preintegration can be implemented directly on top of IMU propagation. 
The propagation error state $\delta\mathbf{x}_{n,e|s}$ need not coincide with the one used by the optimizer. More broadly, improvements in preintegration methods can also benefit propagation; for example, the closed-form preintegration of \cite{eckenhoffClosedformPreintegrationMethods2019} can be used to improve the integration of \eqref{eq:cont-state} in propagation.

\subsection{Propagation by Preintegration}
\label{sec:prop_by_preint}

The reverse direction is also possible: state propagation can be realized using preintegration. 
First, the state value propagation follows directly from $\mbf f(\mbf x_s^p, \Delta \mbf x_{se}^g(\mbf b_s^p))$ \eqref{subeq:prop-by-int-rot}--\eqref{subeq:prop-by-int-b}. 
Second, to obtain the transition matrix, we 
take differentials of $\mbf f(\mbf x_s^p, \Delta \mbf x_{se}^g(\mbf b_s^p))$ yielding
\begin{equation}
\begin{split}
\delta\mathbf{x}_{e|s}^p &= \boldsymbol \Phi_{e|s} \delta \mbf x_s^p + \mbf G_{e|s} \delta \mbf x_{se}^g,\\
\mbf G_{e|s} &= \frac{\partial \delta \mbf f}{\partial \delta \mbf x_{se}^g} = \begin{bmatrix}
\mbf G_{e|s}^{nn} & \mbf 0_{9\times6}\\
\mbf 0_{6\times9} & \frac{\partial \mbf f}{\partial \delta \mbf b_s^g}
\end{bmatrix},\\
\boldsymbol \Phi_{e|s}
&= \frac{\partial \delta \mbf f}{\partial \delta \mbf x_s^p} +
\begin{bmatrix}
\mathbf{0} & \mbf G_{e|s}
\frac{\partial \delta \mbf x_{se}^g}{\partial \delta \mbf b_s^g}
\frac{\partial \delta \mbf b_s^g}{\partial \delta \mbf b_s^p}
\end{bmatrix}\\
&=
\frac{\partial \delta \mbf f}{\partial \delta \mbf x_s^p} +
\begin{bmatrix}
\mathbf{0} & \mbf G_{e|s}^{nn} \mathbf{J}_{b_s}^n \mbf P_b \\
\mbf 0 & \mbf 0_{6\times6}
\end{bmatrix},
\end{split}
\end{equation}
where
\[
\mbf P_b = \frac{\partial \delta \mbf b_s^g}{\partial \delta \mbf b_s^p}, \quad
\frac{\partial \delta \mbf x_{se}^g}{\partial \delta \mbf b_s^g} = \mbf J_{b_s} = \begin{bmatrix}
    \mbf J_{b_s}^n \\ \mbf 0_{6\times6}
\end{bmatrix}.
\]

Third, with the above derivatives $\boldsymbol \Phi_{e|s}$ and $\mbf G_{e|s}$, the preint measurement covariance $\Sigma_{\Delta\mathbf{x}}$ and the covariance of the initial state $\Sigma_s$, the propagated covariance is obtained by
\begin{align}
\Sigma_{e|s}
&=
\boldsymbol{\Phi}_{e|s}
\Sigma_s
\boldsymbol{\Phi}_{e|s}^{\top}
+
\mathbf{G}_{e|s}
\Sigma_{\Delta\mathbf{x}}
\mathbf{G}_{e|s}^{\top}.
\label{eq:prop-cov}
\end{align}

\subsection{Preintegration Factors in Optimization}
The preint measurement's primary purpose is to build a residual in nonlinear optimization.
A typical residual as realized by \texttt{PreintegrationBase::computeError} in \texttt{GTSAM}, is defined by
\begin{align}
\mathbf{r}_{\theta} &= \Log \left(\mathbf{R}_{we}^{\top} \mathbf{R}_{w, e|s}\right),
\label{subeq:rtheta} \\
\mathbf{r}_{p} &= \mathbf{R}_{we}^{\top} \big(\mathbf{p}_{e|s} - \mathbf{p}_{e}\big), \\
\mathbf{r}_{v} &= \mathbf{R}_{we}^{\top} \big(\mathbf{v}_{e|s} - \mathbf{v}_{e}\big), \\
\mathbf{r}_b &= \mbf b_{e|s} - \mbf b_e.
\label{subeq:rb}
\end{align}
Alternative residual definitions are also possible.
An often desirable synergy between the residual and preint perturbation is that the residual Jacobian relative to the perturbation is close to the identity matrix (see \eqref{eq:forster-res-jac} for example).

To build a factor in graph-based optimization, we need the residual Jacobians and covariance. 
First, we observe that the Jacobians of the residuals with respect to position, velocity, and orientation are independent of the specific preintegration formulas in \eqref{subeq:preint-theta}--\eqref{subeq:preint-t}. 
Therefore, changing the error-state definition only requires re-deriving these Jacobians from the residual definitions, \eg \eqref{subeq:rtheta}--\eqref{subeq:rb}.
Second, the Jacobians of the residuals with respect to the initial bias can be obtained through preintegration:
\begin{equation}
\frac{\partial \mathbf{r}}{\partial \delta\mathbf{b}_s}
= \frac{\partial \mathbf{r}}{\partial \delta\mbf{x}_{se}}
\frac{\partial \delta\mbf x_{se}}{\partial \delta\mathbf{b}_s}.
\end{equation}

The covariance of the residual is obtained by the chain rule:
\begin{align}
\Sigma_{r} =
\frac{\partial \mathbf{r}}{\partial \delta\mbf{x}_{se}}
\ \mathrm{cov}(\delta \mbf x_{se})\
\left(\frac{\partial \mathbf{r}}{\partial \delta\mbf{x}_{se}}\right)^{\top}.
\label{eq:residual-cov}
\end{align}
If the residual perturbation is defined such that
$\frac{\partial \mathbf{r}}{\partial \delta\mbf{x}_{se}} \approx \mbf I$,
then the residual covariance is approximately the preintegrated measurement covariance.

To avoid recomputing preintegration whenever the bias estimate changes during optimization, the residuals are often linearized at a nominal bias $\widehat{\mathbf{b}}$:
\begin{align}
\mathbf{r}_{\theta}
&= \mathbf{r}_{\theta}(\widehat{\mathbf{b}})
+ \frac{\partial \mathbf{r}_{\theta}}{\partial \delta\mathbf{b}_g}\,\delta\mathbf{b}_g, \\
\mathbf{r}_{p}
&= \mathbf{r}_{p}(\widehat{\mathbf{b}})
+ \frac{\partial \mathbf{r}_{p}}{\partial \delta\mathbf{b}}\,\delta\mathbf{b}, \\
\mathbf{r}_{v}
&= \mathbf{r}_{v}(\widehat{\mathbf{b}})
+ \frac{\partial \mathbf{r}_{v}}{\partial \delta\mathbf{b}}\,\delta\mathbf{b}.
\end{align}

\subsection{Implementation}
The above derivations are generic with respect to the error-state definitions of
$\delta \mathbf{x}_s$, $\delta \mathbf{x}_e$, $\delta \mathbf{x}_{e|s}$, and $\delta \mathbf{x}_{s e}$.
In the following validation of the equivalence between preintegration and propagation, we use the left error states \cite{huaiObservabilityAnalysisKeyframebased2022} for
$\delta \mathbf{x}_s^p$, $\delta \mathbf{x}_e^p$, and $\delta \mathbf{x}_{e|s}^p$, and right error states for
$\delta \mathbf{x}_{s e}^g$.

The left error state is defined by
\begin{align}
\mathbf{p} &= \widehat{\mathbf{p}} + \delta \mathbf{p}, \quad
\mathbf{R} = \Exp(\delta \boldsymbol{\theta})\,\widehat{\mathbf{R}}, \label{subeq:left-dpr}\\
\mathbf{v} &= \widehat{\mathbf{v}} + \delta \mathbf{v}, \quad
\mathbf{b}_g = \widehat{\mathbf{b}}_g + \delta \mathbf{b}_g, \\
\mathbf{b}_a &= \widehat{\mathbf{b}}_a + \delta \mathbf{b}_a. \label{subeq:left-dba}
\end{align}

The right error state as realized in \texttt{gtsam::NavState::retract} is defined by
\begin{align}
\mathbf{R} &= \widehat{\mathbf{R}}\Exp(\delta \boldsymbol{\theta}), \quad
\mathbf{p} = \widehat{\mathbf{p}} + \mbf R \delta \mathbf{p}, \label{subeq:navstate-drp} \\
\mathbf{v} &= \widehat{\mathbf{v}} + \mbf R \delta \mathbf{v}, \quad
\mathbf{b}_a = \widehat{\mathbf{b}}_a + \delta \mathbf{b}_a, \\
\mathbf{b}_g &= \widehat{\mathbf{b}}_g + \delta \mathbf{b}_g. \label{subeq:navstate-dbg}
\end{align}

\paragraph{GTSAM Peculiarities.}
In \texttt{GTSAM}, the Jacobians of manifold and tangent preintegrated measurements adopt the right preint perturbation (\ie, error state) $\delta \mbf x_{se}^j$ defined by
\begin{align}
\Delta \mathbf{R}_{s e}
&= \Delta \widehat{\mathbf{R}}_{s e}\,\Exp(\delta \boldsymbol{\theta}^j_{se}), \label{subeq:preint-error-dr}\\
\Delta \mathbf{p}^{s}_{e}
&= \Delta \widehat{\mathbf{p}}^{s}_{e} + \delta \mathbf{p}^{j}_{se}, \label{subeq:preint-error-dp}\\
\Delta \mathbf{v}^{s}_{e}
&= \Delta \widehat{\mathbf{v}}^{s}_{e} + \delta \mathbf{v}^{j}_{se}, \label{subeq:preint-error-dv}\\
\Delta \mbf b_{se} &= [\Delta \mbf b_a^{\top}, \Delta \mbf b_g^{\top}]^{\top} = - \delta \mbf b.
\label{subeq:preint-error-db}
\end{align}

The first three equations \eqref{subeq:preint-error-dr}--\eqref{subeq:preint-error-dv} follow from the code of \texttt{TEST(ManifoldPreintegration, BiasCorrectionJacobians)} and \texttt{TEST(ImuFactor, BiasCorrectionJacobians)}, where the numerical derivatives are defined on the vector space, \eg,
\begin{equation}
\frac{\partial \Delta \mbf p}{\partial \mbf b}
=\lim_{\delta \mbf b\rightarrow \mbf 0}
\frac{\Delta \mbf p(\mbf b + \delta \mbf b) - \Delta \mbf p(\mbf b)}{\delta \mbf b},
\end{equation}
rather than on the NavState manifold.

The covariance of tangent preintegration also follows this perturbation definition.
As a result of \eqref{subeq:rtheta}-\eqref{subeq:rb} and \eqref{subeq:preint-error-dr}-\eqref{subeq:preint-error-db}, we get the residual Jacobian for the tangent preint measurement,
\begin{equation}
\frac{\partial \mathbf{r}}{\partial \delta\mbf{x}_{se}^j} =
\begin{bmatrix}
\mbf J_r^{-1}(\mbf r_{\theta}) & \mbf 0 & \mbf 0 & \mbf 0 & \mbf 0 \\
\mbf 0 & \mbf M & \mbf 0 & \mbf 0 & \mbf 0 \\
\mbf 0 & \mbf 0 & \mbf M & \mbf 0 & \mbf 0 \\
\mbf 0 & \mbf 0 & \mbf 0 & -\mbf I_3 & \mbf 0 \\
\mbf 0 & \mbf 0 & \mbf 0 & \mbf 0 & -\mbf I_3
\end{bmatrix}
\end{equation}
with $\mbf M = \mbf R_{we}^{\top} \mbf R_{ws}$.
The residual Jacobian is clearly not close to the identity matrix, and this must be accounted for when computing the residual covariance \eqref{eq:residual-cov}.

In contrast, the covariance of manifold preintegration adopts a right perturbation $\delta \mathbf{x}_{se}^c$ defined on the NavState manifold:
\begin{equation}
\begin{split}
\Delta \mathbf{R}_{s e}
&= \Delta \widehat{\mathbf{R}}_{s e}\,\Exp(\delta \boldsymbol{\theta}^c_{se}),\\
\Delta \mathbf{p}^{s}_{e}
&= \Delta \widehat{\mathbf{p}}^{s}_{e} + \Delta \mathbf{R}_{s e} \delta \mathbf{p}^{c}_{se}, \\
\Delta \mathbf{v}^{s}_{e}
&= \Delta \widehat{\mathbf{v}}^{s}_{e} + \Delta \mathbf{R}_{s e} \delta \mathbf{v}^{c}_{se}, \\
\Delta \mbf b_{se} &= [\Delta \mbf b_a^{\top}, \Delta \mbf b_g^{\top}]^{\top} = - \delta \mbf b^c.
\label{eq:residual-error}
\end{split}
\end{equation}
The residual Jacobian for the manifold preintegration is
\begin{equation}
\frac{\partial \mathbf{r}}{\partial \delta\mbf{x}_{se}^c} =
\begin{bmatrix}
\mbf J_r^{-1}(\mbf r_{\theta}) & \mbf 0 & \mbf 0 & \mbf 0 & \mbf 0 \\
\mbf 0 & \mbf M & \mbf 0 & \mbf 0 & \mbf 0 \\
\mbf 0 & \mbf 0 & \mbf M & \mbf 0 & \mbf 0 \\
\mbf 0 & \mbf 0 & \mbf 0 & -\mbf I_3 & \mbf 0 \\
\mbf 0 & \mbf 0 & \mbf 0 & \mbf 0 & -\mbf I_3
\end{bmatrix},
\end{equation}
with $\mbf M = \mbf R_{we}^{\top} \mbf R_{ws} \Delta \mbf R_{se}$, and it is close to the identity matrix in practice, except for the sign of the bias Jacobians.

\paragraph{Forster Preintegration.}
The preint residual in \texttt{GTSAM} is defined differently from that in \cite{forsterOnmanifoldPreintegrationRealtime2017}. Forster \etal define the residual as
\begin{align}
\mathbf{r}_{\theta} &= \Log \left(\mathbf{R}_{w,e|s}^{\top} \mathbf{R}_{we}\right),
\label{subeq:rtheta-forster} \\
\mathbf{r}_{p} &= \mathbf{R}_{ws}^{\top} \big(\mathbf{p}_{e} - \mathbf{p}_{e|s}\big), \\
\mathbf{r}_{v} &= \mathbf{R}_{ws}^{\top} \big(\mathbf{v}_{e} - \mathbf{v}_{e|s}\big), \\
\mathbf{r}_b &= \mbf b_{e} - \mbf b_{e|s},
\label{subeq:rp-forster}
\end{align}
and define the preintegrated perturbation as
\begin{align}
\Delta \mathbf{R}_{s e}
&= \Delta \widehat{\mathbf{R}}_{s e} \Exp(-\delta \boldsymbol{\theta}^f_{se}),\\
\Delta \mathbf{p}^{s}_{e}
&= \Delta \widehat{\mathbf{p}}^{s}_{e} - \delta \mathbf{p}^{f}_{se}, \\
\Delta \mathbf{v}^{s}_{e}
&= \Delta \widehat{\mathbf{v}}^{s}_{e} - \delta \mathbf{v}^{f}_{se}, \\
\Delta \mbf b_{se} & = - \delta \mbf b.
\label{eq:preint-error-forster}
\end{align}
The above residual and preint perturbation lead to a residual Jacobian close to the identity matrix,
\begin{equation}
\frac{\partial \mathbf{r}}{\partial \delta\mbf{x}_{se}^f} =
\begin{bmatrix}
\mbf J_l^{-1}(\mbf r_{\theta}) & \mbf 0 & \mbf 0 & \mbf 0 & \mbf 0 \\
\mbf 0 & \mbf I_3 & \mbf 0 & \mbf 0 & \mbf 0 \\
\mbf 0 & \mbf 0 & \mbf I_3 & \mbf 0 & \mbf 0 \\
\mbf 0 & \mbf 0 & \mbf 0 & \mbf I_3 & \mbf 0 \\
\mbf 0 & \mbf 0 & \mbf 0 & \mbf 0 & \mbf I_3
\end{bmatrix},
\label{eq:forster-res-jac}
\end{equation}
that is, Forster's preint measurement covariance is nearly identical to the residual covariance.
For completeness, we also note that Forster's state perturbation is defined as
\begin{align}
\mathbf{R} &= \widehat{\mathbf{R}} \Exp(\delta \boldsymbol{\theta}), \quad
\mathbf{p} = \widehat{\mathbf{p}} + \mbf R \delta \mathbf{p}, \\
\mathbf{v} &= \widehat{\mathbf{v}} + \delta \mathbf{v}, \quad
\mathbf{b}_a = \widehat{\mathbf{b}}_a + \delta \mathbf{b}_a, \\
\mathbf{b}_g &= \widehat{\mathbf{b}}_g + \delta \mathbf{b}_g.
\end{align}

\paragraph{IMU Propagation in Detail.}
Details of IMU propagation can be found in \cite{jekeliInertialNavigationSystems2001}. For the simple case without Earth rotation or gravity deflection, which is typical for consumer-grade IMUs, the continuous-time state model is
\begin{equation}
\frac{d}{dt}
\begin{bmatrix}
\mathbf{p} \\
\mathbf{R} \\
\mathbf{v} \\
\mathbf{b}_g \\
\mathbf{b}_a
\end{bmatrix}
=
\begin{bmatrix}
\mathbf{v} \\
\mathbf{R}\,[\boldsymbol{\omega}_m - \mathbf{b}_g - \mathbf{n}_g]_\times \\
\mathbf{g} + \mathbf{R}\big(\mathbf{a}_m - \mathbf{b}_a - \mathbf{n}_a\big) \\
\mathbf{n}_{w g} \\
\mathbf{n}_{w a}
\end{bmatrix},
\label{eq:diff-state}
\end{equation}
where $\mathbf{a}_m$ and $\boldsymbol{\omega}_m$ are the measured specific force and angular velocity, and
$\mathbf{n}_{\cdot}$ and $\mathbf{n}_{w\cdot}$ denote white measurement noise and bias random walks.

If Hamilton quaternions $\mathbf{q} = [q_w, q_x, q_y, q_z]^\top$ are used to represent orientation, the rotation dynamics become \cite{solaQuaternionKinematicsErrorstate2017}
\begin{equation}
\begin{aligned}
\dot{\mathbf{q}} &= \tfrac{1}{2}\,\Omega\big(\boldsymbol{\omega}_m - \mathbf{b}_g - \mathbf{n}_g\big)\,\mathbf{q}, \\
\dot{\mathbf{v}} &= \mathbf{g} + \mathbf{R}(\mathbf{q})\big(\mathbf{a}_m - \mathbf{b}_a - \mathbf{n}_a\big),
\end{aligned}
\end{equation}
with
\[
\Omega(\boldsymbol{\omega}) =
\begin{bmatrix}
0 & -\boldsymbol{\omega}^\top \\
\boldsymbol{\omega} & -[\boldsymbol{\omega}]_\times
\end{bmatrix}.
\]

For the left error state, the error dynamics are
\begin{align}
\frac{d\,\delta \boldsymbol{\theta}}{dt} &= \mathbf{R}\,\delta \boldsymbol{\omega}, \\
\frac{d\,\delta \mathbf{v}}{dt} &= \mathbf{R}\,\delta \mathbf{a}
+ \delta \boldsymbol{\theta} \times \big(\mathbf{R}\,\mathbf{a}\big), \\
\frac{d\,\delta \mathbf{p}}{dt} &= \delta \mathbf{v}, \\
\frac{d\,\delta \mathbf{b}_g}{dt} &= \mathbf{n}_{w g}, \qquad
\frac{d\,\delta \mathbf{b}_a}{dt} = \mathbf{n}_{w a},
\end{align}
where
\begin{align}
\delta \mathbf{a}
&= \mathbf{a} - \widehat{\mathbf{a}}
= \big(\mathbf{a}_m - \mathbf{b}_a - \mathbf{n}_a\big)
- \big(\mathbf{a}_m - \widehat{\mathbf{b}}_a\big)
= -\delta \mathbf{b}_a - \mathbf{n}_a, \\
\delta \boldsymbol{\omega}
&= \boldsymbol{\omega} - \widehat{\boldsymbol{\omega}}
= \big(\boldsymbol{\omega}_m - \mathbf{b}_g - \mathbf{n}_g\big)
- \big(\boldsymbol{\omega}_m - \widehat{\mathbf{b}}_g\big)
= -\delta \mathbf{b}_g - \mathbf{n}_g.
\end{align}

\section{Experiments}

We evaluate the two-way equivalence between IMU propagation and IMU preintegration in two sets of synthetic experiments.

First, we test \emph{preintegration by propagation}. We generate random IMU data at $200\,\mathrm{Hz}$ for $10\,\mathrm{s}$ in each trial and repeat the experiment 100 times. We compare the Jacobian of the preintegrated measurement with respect to the initial bias, as well as the preintegrated covariance, against the corresponding outputs from \texttt{GTSAM}. The \texttt{GTSAM} implementation uses a right perturbation \eqref{subeq:navstate-drp}-\eqref{subeq:navstate-dbg} for error state, while our propagation module uses a left perturbation \eqref{subeq:left-dpr}-\eqref{subeq:left-dba} for error state.

For a reference entry $r$ and an estimate $e$, we define the entrywise difference as
\begin{equation}
\mathrm{d} =
\begin{cases}
\displaystyle \left| \frac{r - e}{r} \right|, & \text{if } |r| > 10^{-4}, \\
|r - e|, & \text{otherwise}.
\end{cases}
\end{equation}

Following the derivation in Section~\ref{sec:preint_by_prop}, we convert the outputs of IMU propagation into preintegrated quantities and compare the resulting bias Jacobians and preintegrated covariances with those produced by \texttt{GTSAM}'s \texttt{PreintegratedCombinedMeasurements}, using both \texttt{TangentPreintegration} and \texttt{ManifoldPreintegration}. For each trial, we compare the matrices obtained through propagation and conversion against those returned directly by \texttt{GTSAM}.

\begin{figure}[!htbp]
    \centering
    \includegraphics[width=0.5\linewidth]{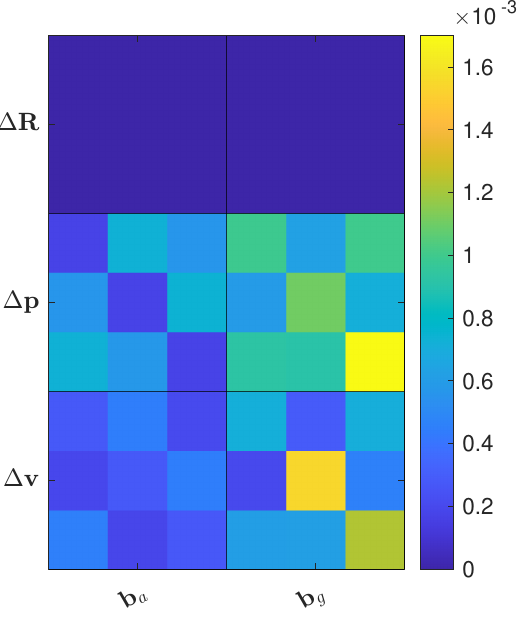} \\
    \includegraphics[width=0.5\linewidth]{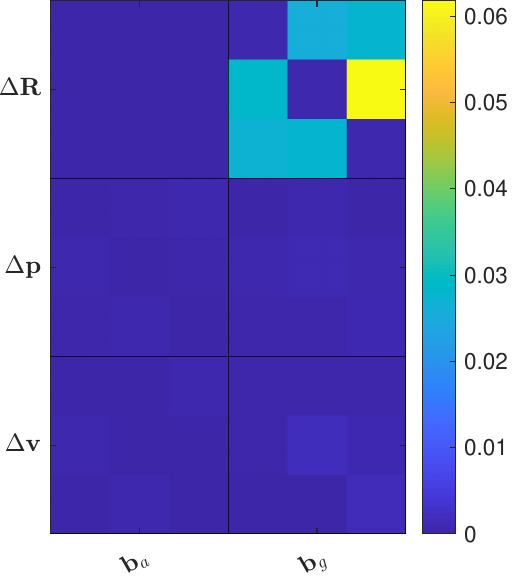} \\
    \includegraphics[width=0.8\linewidth]{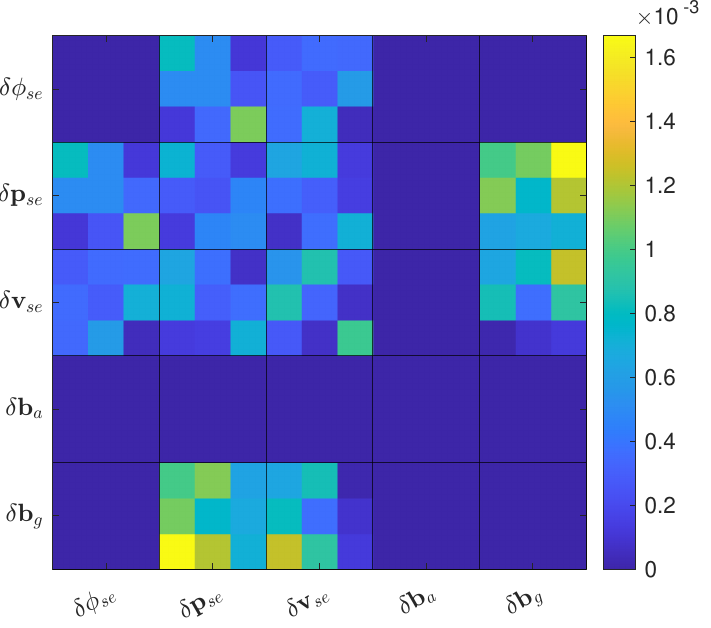} \\
    \includegraphics[width=0.8\linewidth]{figures/heatmaps/preint_by_prop_manifold_Sigma_z_heatmap.pdf} \\
    \caption{Average entrywise differences between the preintegrated bias Jacobians and covariances produced by \texttt{GTSAM} and those obtained by propagation followed by conversion. The top two figures show the bias Jacobian differences for manifold and tangent preintegration, respectively. The bottom two figures show the corresponding covariance differences. In each case, the \texttt{GTSAM} result is used as the reference.}
    \label{fig:preint-diffs}
\end{figure}

The average differences in the bias Jacobians and covariances over 100 runs are shown in Fig.~\ref{fig:preint-diffs}. The close agreement confirms that IMU preintegration can be realized through IMU propagation, provided that the perturbation conventions are handled properly. The results indicate that an existing IMU propagation implementation can be reused to construct preintegration factors for optimization-based multisensor fusion, \eg, radar--inertial mapping.

Next, we test \emph{propagation by preintegration}. Following Section~\ref{sec:prop_by_preint}, we convert the preintegrated measurements into propagated transition matrices and propagated covariances. For each trial, we compute the transition matrix and covariance from \texttt{GTSAM}'s \texttt{PreintegratedCombinedMeasurements}, using either \texttt{TangentPreintegration} or \texttt{ManifoldPreintegration}, and compare them with those produced by our RK4-based propagation implementation initialized from a predefined covariance and a random navigation state.

\begin{figure}[!htbp]
    \centering
    \includegraphics[width=0.85\linewidth]{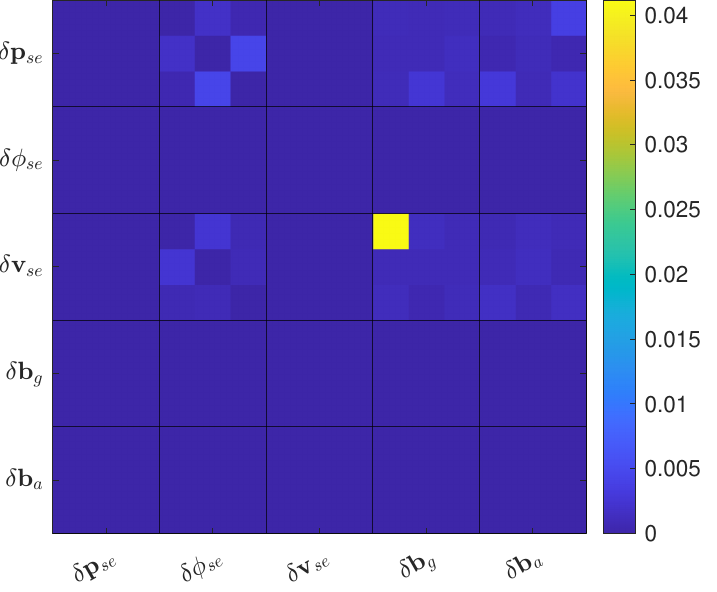} \\
    \includegraphics[width=0.85\linewidth]{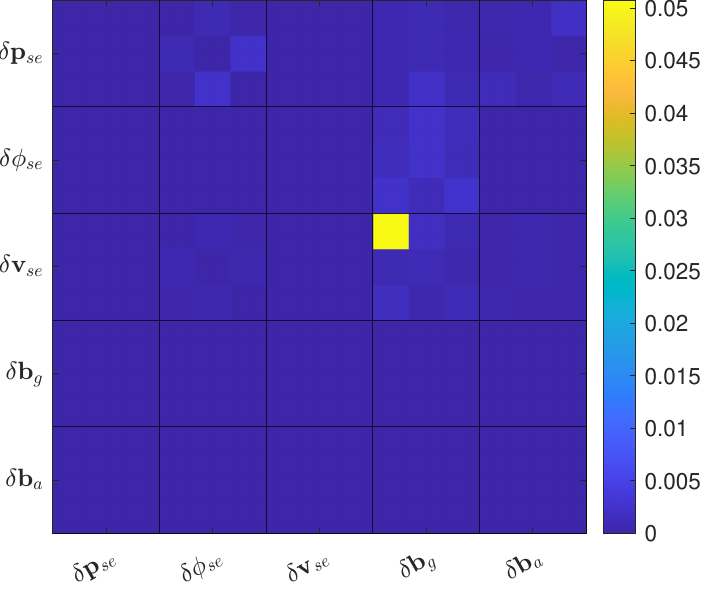} \\
    \includegraphics[width=0.85\linewidth]{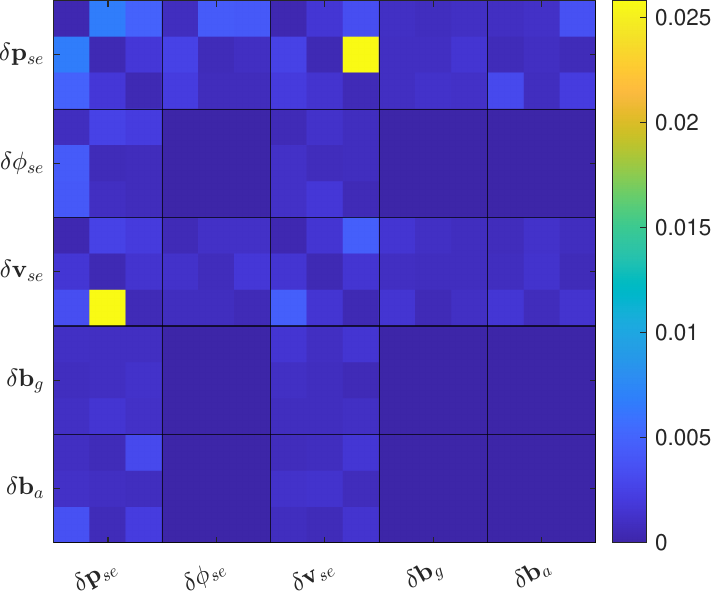} \\
    \includegraphics[width=0.85\linewidth]{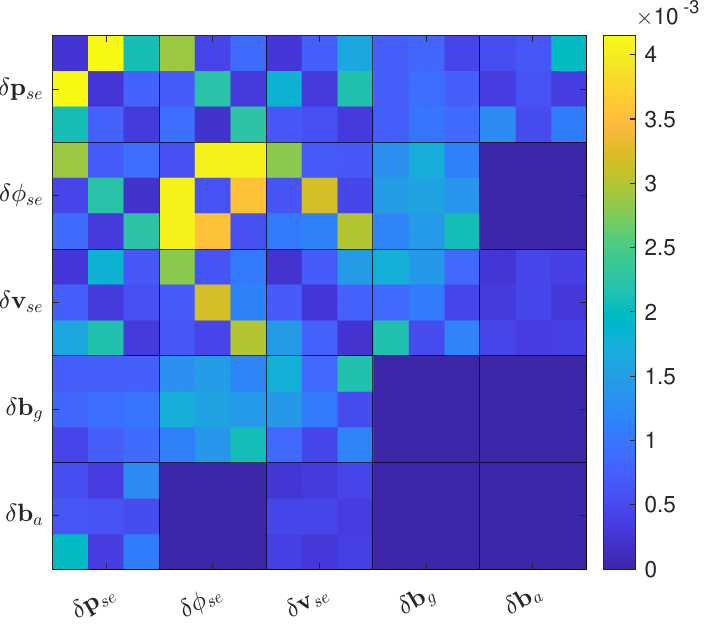} \\
    \caption{Average entrywise differences between the propagated transition matrices and covariances obtained from RK4 propagation with the left error state in \eqref{subeq:left-dpr}-\eqref{subeq:left-dba}, and those recovered from \texttt{GTSAM} preintegration after perturbation conversion. The top two figures correspond to the transition matrix recovered from manifold and tangent preintegration, resp. The bottom two figures show the corresponding covariance differences. The RK4 propagation result is used as the reference.}
    \label{fig:prop-diffs}
\end{figure}

The average differences in the transition matrices and propagated covariances over 100 runs are shown in Fig.~\ref{fig:prop-diffs}. The close agreement confirms that IMU propagation and IMU preintegration encode equivalent information, provided that the perturbation conventions are converted consistently.

Overall, these experiments support the main claim of this note: IMU preintegration can be implemented by wrapping a conventional IMU propagation module, and a preintegration module can in turn be used to recover propagation quantities. This equivalence is useful both for implementation reuse and for validation.

{
    \begin{spacing}{1.17}
        \normalsize
        \bibliography{zotero}
    \end{spacing}
}

\end{document}